%% file: main.tex
\documentclass[10pt,twocolumn,letterpaper]{article}

\usepackage{wacv}              
\definecolor{wacvblue}{rgb}{0.21,0.49,0.74}
\usepackage[pagebackref,breaklinks,colorlinks,allcolors=wacvblue]{hyperref}

\usepackage{graphicx} 
\usepackage{caption}
\usepackage{subcaption}
\usepackage{amsmath}
\usepackage{multicol}
\usepackage{amsfonts}
\usepackage{soul}
\usepackage{algorithm2e}
\usepackage{amssymb}
\usepackage{bbding}
\usepackage{xcolor}
\usepackage{pifont}
\usepackage{xifthen}
\usepackage{multirow}
\usepackage[numbers]{natbib}
\usepackage{tikz}
\usepackage{pgfplots}
\usepackage{scalerel}
\usetikzlibrary{positioning, decorations.pathreplacing,calligraphy, calc, shapes.geometric, intersections, arrows}
\usepgfplotslibrary{fillbetween}
\pgfplotsset{compat=1.18}

\SetKwComment{Comment}{/* }{ */}
\RestyleAlgo{ruled}

\newcommand{\cmark}{\ding{51}}%
\newcommand{\xmark}{\ding{55}}%

\newcommand{\FB}{F'}

\newcommand{\hinput}{\mathbf{v}}
\newcommand{\houtput}{\mathbf{u}}

\title{Deep Feedback Models}

\author{David Calhas\\
INESC-ID\\
Instituto Superior Tecnico\\
{\tt\small david.calhas@tecnico.ulisboa.pt}
\and
Arlindo L. Oliveira\\
INESC-ID\\
Instituto Superior Tecnico\\
{\tt\small arlindo.oliveira@tecnico.ulisboa.pt}
}

\begin{document}

\maketitle




\begin{abstract}
    Deep Feedback Models (DFMs) are a new class of stateful neural networks that combine bottom up input with high level representations over time. This feedback mechanism introduces dynamics into otherwise static architectures, enabling DFMs to iteratively refine their internal state and mimic aspects of biological decision making. We model this process as a differential equation solved through a recurrent neural network, stabilized via exponential decay to ensure convergence. To evaluate their effectiveness, we measure DFMs under two key conditions: robustness to noise and generalization with limited data. In both object recognition and segmentation tasks, DFMs consistently outperform their feedforward counterparts, particularly in low data or high noise regimes. In addition, DFMs translate to medical imaging settings, while being robust against various types of noise corruption. These findings highlight the importance of feedback in achieving stable, robust, and generalizable learning. Code is available at \href{https://github.com/DCalhas/deep_feedback_models}{github.com/DCalhas/deep\_feedback\_models}.
\end{abstract}

\section{Introduction}\label{section:intro}

\input{intro/intro}

\section{Problem description}\label{section:problem}

\input{problem/problem}

\section{Feedback}\label{section:methods}

\input{methods/methods}

\section{Experimental setting}\label{section:experimental}

\input{results/noise/figures/noise_figures}

\input{setting/setting}

\section{Results}\label{section:results}

\input{results/results}

\section{Discussion}\label{section:discussion}

\input{discussion/discussion}

\section{Related work}\label{section:related}

\input{related_work/related_work}
\section{Conclusion}\label{section:conclusion}

\input{conclusion/conclusion}

\input{acknowledgments}

\bibliographystyle{unsrtnat}
\bibliography{bibliography}

\end{document}

%% file: intro/intro.tex
Although artificial neural networks draw inspiration from biology, they are typically static once a prediction is made, unable to revisit or refine their output based on subsequent processing. This contrasts with the dynamic and adaptive nature of the human brain, which continually updates its internal model of the world to minimize the mismatch between expectations and sensory data. The predictive coding framework offers a computational account of this process {\cite{rao1999predictive,clark2024experience}}, positing that the brain operates as a hierarchical inference system. In this view, higher cortical regions generate anticipations about sensory input, while lower regions compute the discrepancy, or prediction error, between those expectations and the actual signal. These error signals drive updates to the internal state, allowing perception to evolve iteratively. Inspired by this model, we propose a feedback based neural architecture that integrates prediction error into its internal state over time. By recurrently incorporating its own output into subsequent computations, the model refines its internal representations through multiple cycles of inference. This iterative scheme contrasts with conventional feedforward networks, which operate in a single computational pass. Our approach mimics perceptual inference in the brain, where interpretation emerges gradually by reducing the gap between top-down expectations and bottom-up evidence {\cite{rao1999predictive,alamia2023role}}. In particular, this inference process does not follow a fixed path, but instead adapts dynamically to the input, enabling convergence to a coherent perceptual state {\cite{bogacz2017tutorial}}.

We introduce \textit{Deep Feedback Models} (DFMs), a new class of neural architectures in which high-level predictions are fed back into earlier processing stages, allowing internal states to evolve dynamically over time. DFMs are inspired by a core principle in neuroscience, \textit{predictive coding}, which posits that the brain continually refines its internal representations by minimizing the mismatch between expected and actual sensory inputs \cite{rao1999predictive, friston2005theory}.\footnote{A video demonstration of our code is available at \href{https://youtu.be/SNCrZqiDu5g}{https://youtu.be/SNCrZqiDu5g}} Unlike prior recurrent models that repeat layers or use shallow feedback in the final stages \cite{spoerer2017recurrent,linsley2005stable, goetschalckx2023computing}, DFMs implement full network feedback loops: the evolving state is propagated through the network and recursively informs future updates, as illustrated in Figure \ref{fig:graphical_abstract}. This mechanism enables internal representations to be refined toward consistent predictions, an essential characteristic of predictive coding. In this work, we evaluate DFMs in tasks requiring robustness and generalization, and report several key findings:
\begin{itemize}
    \item DFMs consistently outperform feedforward baselines under increasing levels of input noise (Section \ref{section:robustness});
    \item DFMs generalize from fewer labeled examples, achieving accurate predictions in few-shot settings (Section \ref{section:generalization});
    \item DFMs translate to medical imaging applications, which are characterized by limited observations and high levels of noise, outperforming their feedforward counterparts in these settings (Section \ref{section:medical_application});
    \item DFMs consistently outperform feedforward baselines in different types of noise (Section \ref{section:medical_application}).
\end{itemize}

Our contributions are as follows:
\begin{itemize}
    \item we propose a biologically inspired feedback architecture (Section \ref{section:methods}) that enables stable learning on large scale data through an orthogonalization step (Gram-Schmidt, see Section \ref{section:orthogonality});
    \item we extend this framework to image segmentation by introducing a spatially dissipative exponential decay layer (Section \ref{section:conv_exp_decay});
    \item we present extensive experiments comparing DFMs and their feedforward counterparts across noise, data efficiency, and medical imaging settings (Section \ref{section:experimental});
    \item we offer an analysis of the internal dynamics and stability of DFMs (Section \ref{section:discussion}), and position our work within the broader context of biologically inspired and equilibrium based models (Section \ref{section:related}).
\end{itemize}

%% file: problem/problem.tex
Traditional artificial neural networks are \textit{static} systems: given an input $\mathbf{x} \in \mathbb{R}^{C \times H \times W}$ and parameters $\theta$, the network, $F: \mathbb{R}^{C \times H \times W} \to \mathbb{R}^{N \times h \times w}$, produces a fixed internal signal $F(\mathbf{x}; \theta)$ and a final prediction $G(F(\mathbf{x}; \theta))$, where $G: \mathbb{R}^{N \times h \times w} \to \mathbb{R}^{L}$ making $G \circ F: \mathbb{R}^{C \times H \times W} \to \mathbb{R}^{L}$. Naturally, $H \times W$ represents the height and width of an image, whereas $h \times w$ represents height and width of a latent representation. Despite efforts to make artificial networks more biologically plausible \cite{kubilius2018cornet,dapello2020simulating,rajalingham2022recurrent}, most architectures lack one crucial feature of biological systems, \textit{feedback}, the ability for high level predictions to recursively influence earlier layers. To incorporate feedback, we introduce a dynamical system with a time dependent state vector $\mathbf{h}(t) = [\hinput(t), \houtput(t)] \in \mathbb{R}^{(B+N) \times H \times W}$, which is iteratively updated by a deep neural function $\FB: \mathbb{R}^{(C+B) \times H \times W} \to \mathbb{R}^{(N+B) \times h \times w}$, similar to $F$ but the final representation is upsampled. The state is split into a feedback component $\hinput(t) \in \mathbb{R}^{B \times h \times w}$, which is softmax normalized and concatenated with the input, and an output component $\houtput(t) \in \mathbb{R}^{N \times h \times w}$, which is decoded by a classifier $G$:
\begin{align}\label{equation:feedback}
    \delta(\mathbf{x}, \mathbf{h}, t) &= \FB \left(\left[ \mathbf{x}, \frac{\exp\left(\hinput(t)\right)}{\sum_i \exp\left(\hinput_i(t)\right)} \right]; \theta \right) \\ \hat{\mathbf{y}}(t) &= G\left( \houtput(t) \right).
\end{align}

Let $\mathcal{D} = \bigcup_i \{ \mathbf{x}_i, \mathbf{y}_i \}$ be a dataset where each input $\mathbf{x}_i \in \mathbb{R}^{C \times H \times W}$ is an image, and the corresponding ground truth $\mathbf{y}_i$ is either a label vector $\in \mathbb{R}^L$ (for classification tasks) or a segmentation mask $\in \mathbb{R}^{L \times H \times W}$. Here, $C$ is the number of input channels, $H \times W$ is the spatial resolution, and $L$ is the number of classes. Our objective is to learn model parameters, $\theta$, by minimizing the cross entropy loss over the dataset as
\begin{align}\label{equation:cross_entropy}
    &\mathcal{L}(\theta) = \nonumber\\ &- \sum_i^{|\mathcal{D}|} \mathbf{y}_i 
    \text{log} \circ G \circ \FB \left(\left[ \mathbf{x}_i, \frac{\exp\left(\hinput(T)\right)}{\sum_j \exp\left(\hinput_j(T)\right)} \right]; \theta \right) .
\end{align}
We evaluate two key capabilities of the model: robustness to noise and generalization from few examples. These are tested under the following two experimental settings:
\begin{enumerate}
    \item \textbf{Noise setting}: Gaussian noise $\epsilon \sim \mathcal{N}(0, \sigma^2)$ is added to each input, i.e., $\mathbf{x}_i + \epsilon$, for a fixed dataset size $|\mathcal{D}|$. This simulates degradation in sensory input and challenges the model's ability to extract reliable features under corruption.
    \item \textbf{Few-shot setting}: The number of labeled examples per class is limited to $D \ll |\mathcal{D}| / L$, with no added noise (i.e., $\sigma = 0$). We sample $D$ instances per class uniformly from the dataset $\mathcal{D}$. This setting tests the model’s ability to generalize from limited supervision.
\end{enumerate}
These two conditions are mutually exclusive: noise is applied only when all training data are used, and few-shot generalization is evaluated on clean data. Formally, the combined training objective in either case becomes
\begin{align}\label{equation:cross_entropy_noise_examples}
    &\mathcal{L}(\theta) = \nonumber \\ &- \sum_i^{D \times L} \mathbf{y}_i \text{log} \circ G \circ \FB \left(\left[ \mathbf{x}_i+\epsilon, \frac{\exp\left(\hinput(T)\right)}{\sum_j \exp\left(\hinput_j(T)\right)} \right]; \theta \right),
\end{align}
where $\epsilon = 0$ in the few-shot case, and $D \times L = |\mathcal{D}|$ in the noisy case. We perform few-shot experiments only for classification tasks, since segmentation requires nontrivial control over per-pixel class distributions. Throughout, we apply this setup to evaluate both feedforward and DFMs under controlled perturbations and limited supervision, in order to measure their relative strengths.

%% file: methods/methods.tex
\input{intro/figures/graphical_abstract}
\input{results/figures/path_two_classes}
DFMs operate as dynamical systems, where internal representations evolve over time. However, in contrast to classical recurrent cells or shallow feedback loops \cite{spoerer2017recurrent, linsley2005stable, goetschalckx2023computing}, we apply feedback across a full deep neural network, as illustrated in Figure \ref{fig:graphical_abstract}. This introduces significant challenges because deep architectures are highly non linear and difficult to stabilize, which makes ensuring convergence or contractivity of the underlying dynamics non trivial, especially in large-scale settings. To guarantee that the system stabilizes, i.e. the internal state $\mathbf{h}(t)$ does not diverge, we introduce an exponential decay mechanism defined as
\begin{equation}\label{equation:feedback_decay}
    \mathbf{h}(T) = \mathbf{h}(0) + \sum_{t=0}^{T-1}\delta(\mathbf{h}, t)^\top \cdot Q \cdot e^{\tau^{-1} \cdot t \cdot  \Sigma} \cdot Q^{-1},
\end{equation}
where $\delta(\mathbf{h}, t)$ is the signal produced by $\FB$ defined in equation \ref{equation:feedback}, $Q$ represents the matrix with eigenvectors in the columns, and $\Sigma$ are the eigenvalues of the system. In order for the system to decay over time (see Figure \ref{fig:fb_path}), we define $\Sigma=-I$. Note that the projection matrix is $P = Q \cdot  \Sigma \cdot Q^{-1}$, and its exponential becomes trivial, $e^P = Q \cdot e^\Sigma \cdot Q^{-1}$. This matrix defines the iterative feedback dynamics of the neural network as $\frac{\mbox{d}\mathbf{h}}{\mbox{d} t} = \delta(\mathbf{h},t)^{\top} \cdot  e^{\tau^{-1} \cdot t \cdot  P}$. In practice we use a forward-Euler discretization ($\Delta t = 1$), producing the discrete update shown in equation \ref{equation:feedback_decay}. This decay mimics a dissipative process, gradually damping out feedback energy over time. Biologically, this reflects the natural temporal smoothing in cortical circuits \cite{gros2007neural}; mathematically, it introduces an attractor like behavior that leads to asymptotic stability.

\subsection{Convolution with exponential decay}\label{section:conv_exp_decay}

While the exponential decay mechanism described above enforces temporal stability by damping feedback signals, it applies uniformly across spatial locations, treating each pixel independently. However, in dense prediction tasks like semantic segmentation, spatial interactions between neighboring pixels are essential for accurate boundary localization and region coherence. A convolutional layer is parametrized by $\mathbf{w} \in \mathbb{R}^{k_h \times k_w \times C_{in} \times C_{out}}$. This operation can be modeled for exponential decay taking into account the spatial dimension considered in a convolution. Consider a single kernel window associated to an input and output channel, $j$, such that $\mathbf{w}^{j} \in \mathbb{R}^{k_h \times k_h}$ is square, i.e. $k_h=k_w$, then $\mathbf{w}^{j} = Q_{j} \cdot \Sigma_{j} \cdot Q^{-1}_{j}$ is the eigendecomposition of the kernel window where columns of $Q$ are eigenvectors and the diagonal of $\Sigma$ contains the eigenvalues. 
For simplicity, we refer to a convolution simply as $\mathbf{x} \ast \mathbf{w}$. We can now define a system that evolves in time by doing spatial exponential decay as 
\begin{equation}\label{equation:ode_conv}
    \frac{\mbox{d}\mathbf{h}}{\mbox{d} t} = \delta(\mathbf{h}, t) \ast e^{\tau^{-1}\cdot t\cdot \mathbf{w}},
\end{equation}
whose Euler discretization is defined by 
\begin{equation}\label{equation:ode_conv_solution}
    \mathbf{h}(T) = \mathbf{h}(0) + \sum_{t=0}^{T-1} \delta(\mathbf{h}, t) \ast e^{\tau^{-1}\cdot t\cdot \mathbf{w}}.
\end{equation}
This system naturally dissipates pixel's magnitude in the spatial dimensions defined by the kernel and is easily solved by exponentiating the eigenvalues as
\begin{equation}\label{equation:ode_conv_exp_eig}
    \mathbf{h}(T) = \mathbf{h}(0) + \sum_{t=0}^{T-1} \delta(\mathbf{h}, t) \ast \left( Q \cdot e^{\tau^{-1}\cdot t\cdot \Sigma} \cdot Q^{-1} \right),
\end{equation}
whose limit is naturally $\lim_{t\to \infty} \mathbf{h}(t+1) - \mathbf{h}(t) = 0$.

\textbf{Normalization.} In the backward pass, the gradient flows through a pixel of the input image $k_h^2$ times, corresponding to the total number of pixels in the kernel window. This gradient accumulation is normalized by a constant $Z=\frac{1}{k_h}$, such that $h(t+1)= h(t)+\frac{1}{Z} \frac{\mbox{d}\mathbf{h}}{\mbox{d}t}$.

\subsection{Orthogonality}\label{section:orthogonality}

Until now, we are considering a recurrent system, whose Jacobian is defined as
\begin{equation}\label{equation:jacobian}
    \mathbf{J}_{\mathbf{h}} = 1 + \frac{\partial \delta(\mathbf{h}, t)}{\partial \mathbf{h}}  \ast \left[  Q \cdot e^{\tau^{-1}\cdot t\cdot \Sigma} \cdot Q^{-1} \right].
\end{equation} 
The fact that the Jacobian contains a neural network with a high level of nonlinearities makes this problem difficult and more susceptible to the nature of $\FB$, which may result in vanishing or exploding gradients. Jacobian regularizations \cite{bai2019deep, goetschalckx2023computing, linsley2005stable} are suitable when $\FB$ is a simple cell. However when it contains a high level of complex nonlinearities, it becomes unfeasible to apply them. In equation \ref{equation:jacobian}, the exponential decay already guarantees a stable property for the system. If we wish to train $Q$ and $Q^{-1}$, these can neither amplify nor shrink the gradient/response. For this, we introduce an orthogonal correction step after each gradient update. This goes in accordance with Riemann optimization \cite{becigneul2018riemannian}, but instead of doing Householder projections (is computationally expensive and therefore does not allow training in large scale data), we perform a simple QR decomposition, a.k.a. Gram-Schmidt algorithm \cite{leon2013gram}, defined as
\begin{equation}\label{equation:qr}
    U(s) = Q(s) \cdot R(s), 
\end{equation}
where $U$ is the non orthogonal representation of $Q$, $Q$ is an orthogonal matrix, $R$ is an upper triangular matrix, and $s$ denotes the $s$th optimization step. 
Every time there is a gradient update to $Q$, as
\begin{equation}\label{equation:update}
    U(s) = Q(s-1) + \eta \nabla_{Q(s-1)} \mathcal{L},
\end{equation}
equation \ref{equation:qr} is applied to correct $U$ to $Q$, allowing us to write $Q^{-1}= Q^\top$. This ensures stability in training and enables the model to learn under super fast convergence settings \cite{smith2019super}, which would not be possible otherwise because of numerical underflow or overflow. %
We show the training scheme for the proposed methodology in Algorithm \ref{algorithm:training}.

\input{methods/algorithm}

%% file: intro/figures/graphical_abstract.tex
\begin{figure}[ht]
    \centering
    \begin{tikzpicture}
        \node (input) at (0,0) {$\mathbf{x}$};
        \node (F) [draw, rectangle, minimum width=30, minimum height=30, inner sep=0.0, line width=1.5] at ($(input)+(0.25,1.75)$) {$F'$};

        \node (delta) [anchor=west, draw, black!30, rectangle, minimum height=15, minimum width=15, line width=1] at ($(F.east)+(0.4,0)$) {\textcolor{black}{$\delta(\mathbf{h}, t)$}};

        \node (expdecay) [anchor=west, draw, rectangle, minimum height=30, minimum width=15, line width=1.5] at ($(delta.east)+(0.4,0)$) {$e^{\tau^{-1}\cdot t\cdot P}$};
        \node (sum) [anchor=west, draw, circle, minimum size=15, inner sep=0.0] at ($(delta.east)+(2.25,0)$) {$+$};

        \node (h) [draw, black!30, rectangle, minimum height=15, minimum width=30, line width=1] at ($(sum)+(1.2,-0.75)$) {\textcolor{black}{$\mathbf{h}$}};

        \node (v) [draw, black!30, rectangle, minimum height=15, minimum width=15, line width=1] at ($(h)+(-1.4,-1)$) {\textcolor{black}{$\mathbf{v}$}};
        \node (u) [draw, black!30, rectangle, minimum height=15, minimum width=15, line width=1] at ($(h)+(1.4,-1)$) {\textcolor{black}{$\mathbf{u}$}};

        \node (upsample) [anchor=east, draw, rectangle, minimum width=50, minimum height=40, inner sep=0.0, line width=1.5, align=center] at ($(v.west)+(-1.1,0.)$) {\small upsample \\ \small \& \\ \small softmax \normalsize};
        \node (split) [anchor=center] at ($(h)+(0.,-1.3)$) {vector split};

        \node (G) [draw, rectangle, minimum height=20, minimum width=20, line width=1.5] at ($(u)+(0,1.1)$) {$G$};

        \node (output) at ($(G.north)+(0,0.7)$) {$\hat{\mathbf{y}}$};

        \draw[->, line width=1.5] ($(input.north)+(0,0.1)$) -- ($(F.south)+(-0.25,-0.1)$);
        \draw[->, line width=1.5] ($(F.east)+(0.1,0)$)--($(delta.west)+(-0.1,0)$);
        \draw[->, line width=1.5] ($(delta.east)+(0.1,0)$)--($(expdecay.west)+(-0.1,0)$);
        \draw[->, line width=1.5] ($(expdecay.east)+(0.1,0)$)--($(sum.west)+(-0.1,0)$);

        \draw[->, line width=1.5] ($(sum.east)+(0.1,0)$) -- ($(h)+(0,0.75)$) -- ($(h.north)+(0,0.1)$);
        \draw[->, line width=1.5] ($(h.west)+(-0.1,0)$) -- ($(sum)+(0,-0.75)$) -- ($(sum.south)+(0,-0.1)$);

        \draw[->, line width=1.5] ($(h.south)+(0,-0.1)$) -- ($(v)+(1.4,0)$) -- ($(v.east)+(0.1,0)$);
        \draw[->, line width=1.5] ($(u)+(-1.4,0)$) -- ($(u.west)+(-0.1,0)$);

        \draw[->, line width=1.5] ($(v.west)+(-0.1,0)$) -- ($(upsample.east)+(0.1,0)$);
        \draw[->, line width=1.5] ($(upsample.west)+(-0.1,0)$) -- ($(F)+(0.25,-1.75)$) -- ($(F.south)+(0.25,-0.1)$);

        \draw[->, line width=1.5] ($(u.north)+(0,0.1)$)--($(G.south)+(0,-0.1)$);
        \draw[->, line width=1.5] ($(G.north)+(0,0.1)$)--($(output.south)+(0,-0.1)$);
    \end{tikzpicture}
    \caption{Neural network circuit with high level feedback.}
    \label{fig:graphical_abstract}
\end{figure}
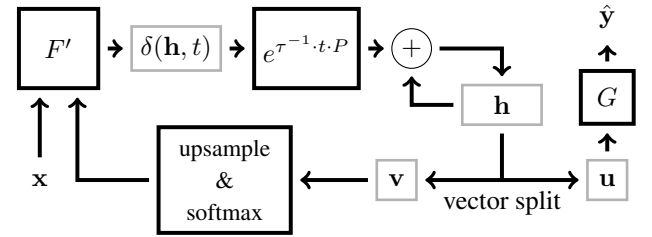

%% file: results/figures/path_two_classes.tex
\begin{figure*}[ht]
    \centering
    \begin{tikzpicture}
        \node (fig) at (0,0) {\includegraphics[width=0.5\textwidth, clip, trim={0.6cm 0.5cm 0.25cm 1.2cm}]{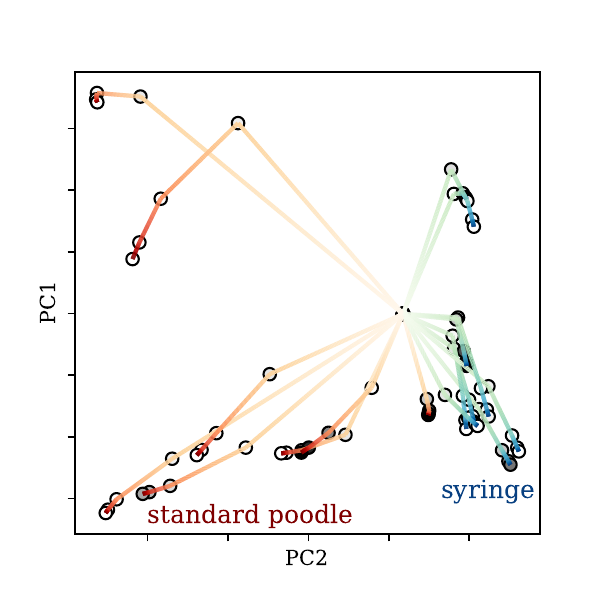}};

        \node (legendsyringe1) [anchor=west] at ($(fig.north east)+(-0.5,-0.75)$) {\includegraphics[width=0.5\textwidth]{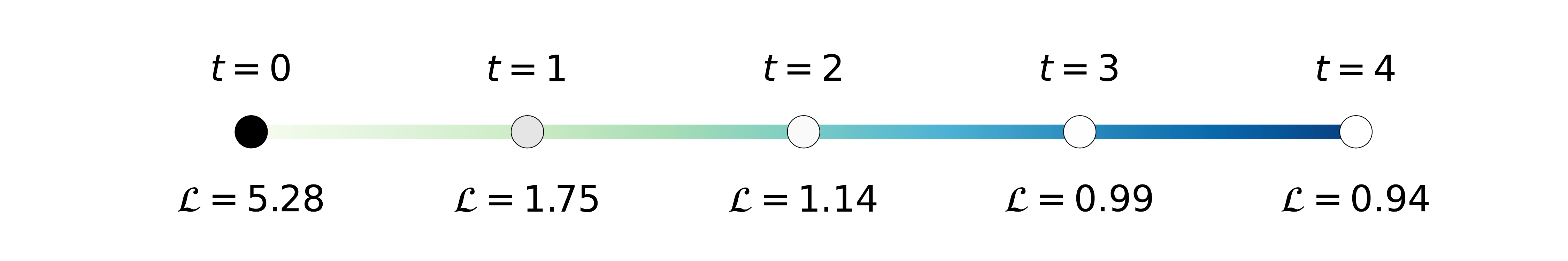}};
        \node (legendsyringe2) at ($(legendsyringe1.south)+(0.,-1.2)$) {\includegraphics[width=0.5\textwidth]{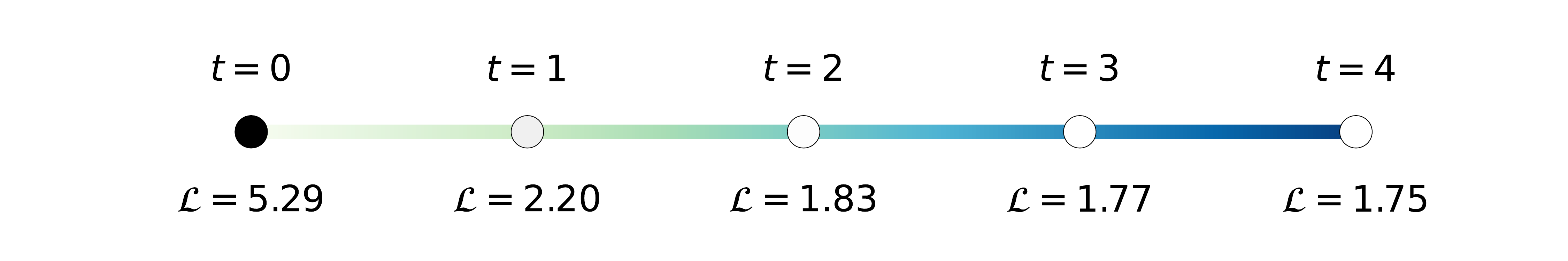}};
        \node (legendpoodle3) at ($(legendsyringe2.south)+(0.,-1.1)$) {\includegraphics[width=0.5\textwidth]{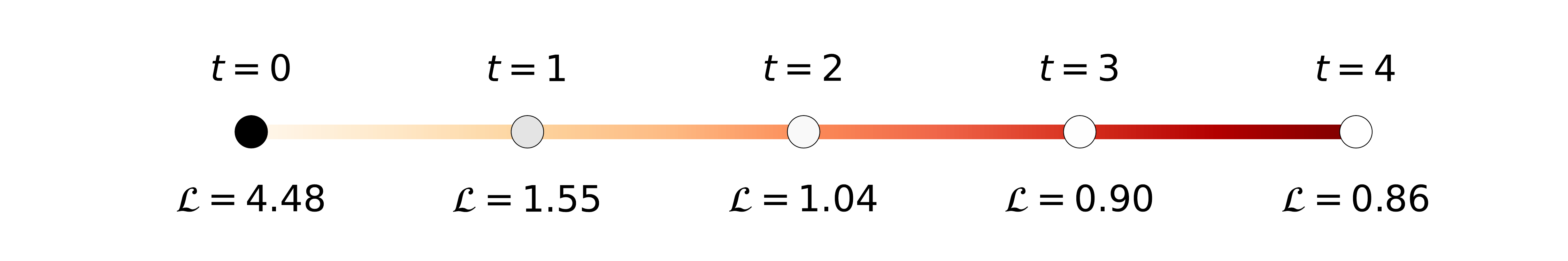}};
        \node (legendpoodle4) at ($(legendpoodle3.south)+(0.,-1.2)$) {\includegraphics[width=0.5\textwidth]{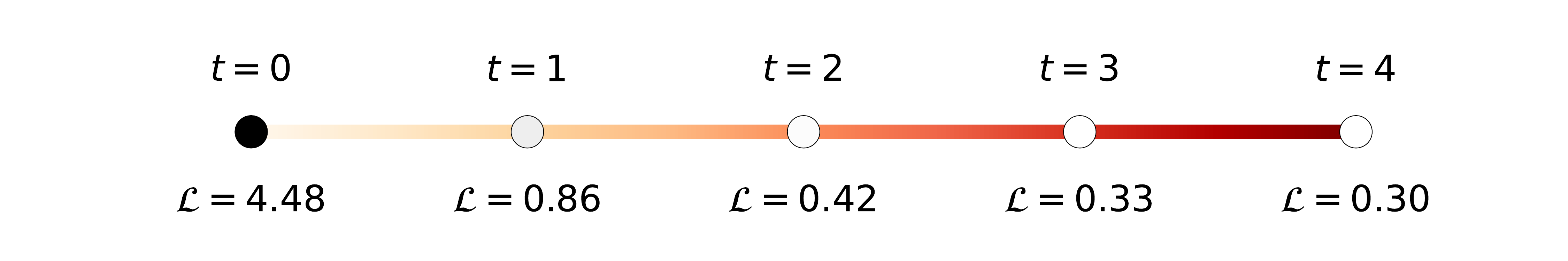}};

        \node (syringerect) [draw, rectangle, line width=0.7, minimum width=210, minimum height=94, anchor=north west] at ($(legendsyringe1.north west)+(0.9,-0.25)$) {};
        \node (poodlerect) [draw, rectangle, line width=0.7, minimum width=210, minimum height=94, anchor=north west] at ($(legendpoodle3.north west)+(0.9,-0.25)$) {};

        \node (syringelegend) [align=center, rotate=90] at ($(syringerect.west)+(-0.3,0)$) {syringe};
        \node (poodlelegend) [align=center, rotate=90] at ($(poodlerect.west)+(-0.3,0)$) {standard poodle};

        \node (legendA) at ($(fig.north west)+(0.3,0)$) {\sffamily{\textbf{A}}};
        \node (legendB) at ($(syringerect.north west)+(-0.4,0.15)$) {\sffamily{\textbf{B}}};
        
    \end{tikzpicture}
    \caption{{\sffamily{\textbf{A}}} illustrates the stability of the proposed system over time. Each line represents the trajectory of a single instance vector $\text{softmax}\left(G(v(t))\right)$. Red lines represent instances of the class \textit{standard poodle}, which is a dog breed, and blue lines represent the class \textit{syringe}. The trajectories are projected onto its two principal components using PCA. To compute this projection, we collect all vectors $\text{softmax}\left(G(v(t))\right)$ from all validation instances of the TinyImageNet dataset \cite{le2015tiny} at each timestep $t$, and perform PCA on this collection to obtain global directions of maximal variance. The same transformation is applied to all trajectories for consistency. The subplots represent the evolution from timestep $t=0$ to $t=4$. The corresponding loss value $\mathcal{L}$ is represented by how dark a marker is at that timestep, the darker the higher the loss. Early in training ($t=0$), the trajectories exhibit large step changes, as the system begins from random initializations $\mathbf{h}(0) \sim \mathcal{N}(0, 0.001)$. {\sffamily{\textbf{B}}} shows the specific loss values at each time steps for two examples of each class each.} 
    \label{fig:fb_path}
\end{figure*}
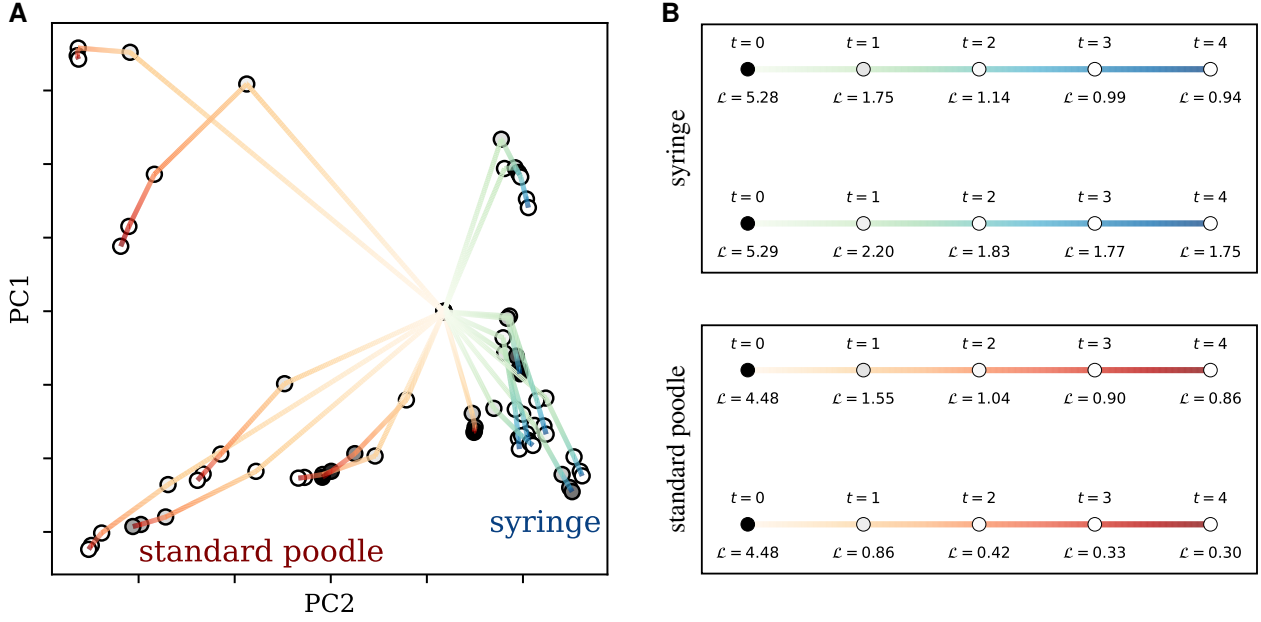

%% file: methods/algorithm.tex
\begin{algorithm}
    \caption{Training scheme for a DFM applying equation \ref{equation:feedback_decay}}\label{algorithm:training}
    $\FB$ \Comment*[r]{a neural network}
    $G$ \Comment*[r]{a classification layer}
    $Q \in \text{O}(B)$ \Comment*[r]{$\mbox{O}(B)$ the set of all orthogonal matrices}
    $\sigma \in \mathbb{R}^{C\times H \times W}$\;
    $D\in \mathbb{N}^+$\;
    $\mathcal{D} = \bigcup_i^{D\times L} \{ \mathbf{x}_i, \mathbf{y}_i \} $\;
    
    \For{$i\gets0$ \KwTo $D\times L$}{
        $\epsilon \sim \mathcal{N}(0, \sigma)$\;
        $\mathbf{x}_i = \mathbf{x}_i + \epsilon$\;
        $\mathbf{u} \sim \mathcal{N}(\mathbf{0}, 0.001)$\;
        $\mathbf{v} \sim \mathcal{N}(\mathbf{0}, 0.001)$\;

        $\mathbf{h} := [ \mathbf{u}, \mathbf{v} ]$ \Comment*[r]{concatenation}
        \For{$t\gets0$ \KwTo $T$}{
            $\delta(\mathbf{h}, t) = \FB \left(\left[ \mathbf{x}, \frac{\mbox{exp}\left(\hinput(t)\right)}{\sum_i \mbox{exp}\left(\hinput_i(t)\right)} \right]; \theta \right)$\;
            $\mathbf{h}(t+1) = \mathbf{h}(t) + \delta(\mathbf{h}, t)^\top \cdot Q \cdot e^{\tau^{-1} \cdot t \cdot \Sigma} \cdot Q^{\top}$\;
            $\hat{\mathbf{y}}(t) = G(\houtput(t))$\;
        }
        $\theta = \theta - \eta \nabla_\theta \mathcal{L}\left(\hat{\mathbf{y}}(T), \mathbf{y}_i\right)$\;
        $U = Q - \eta \nabla_Q \mathcal{L}\left(\hat{\mathbf{y}}(T), \mathbf{y}_i\right)$\;
        $U = \tilde{Q} \cdot R$ \Comment*[r]{QR decomposition}
        $Q=\tilde{Q}$\;
    }
\end{algorithm}

%% file: results/noise/figures/noise_figures.tex
\newcommand{\figheight}{5.25cm}
\newlength{\mscocolegendXshift}
\newlength{\imagenetlegendXshiftL}
\newlength{\imagenetlegendXshiftR}
\pgfmathsetlength{\mscocolegendXshift}{0.63*\figheight}
\pgfmathsetlength{\imagenetlegendXshiftR}{0.63*\figheight}
\pgfmathsetlength{\imagenetlegendXshiftL}{0.63*\figheight}

\begin{figure*}[t]
    \centering
    \begin{tikzpicture}
        \node (plot) at (0cm,0cm) {\includegraphics[clip, trim={3.cm 0.5cm 12cm 1.5cm}, height=\figheight]{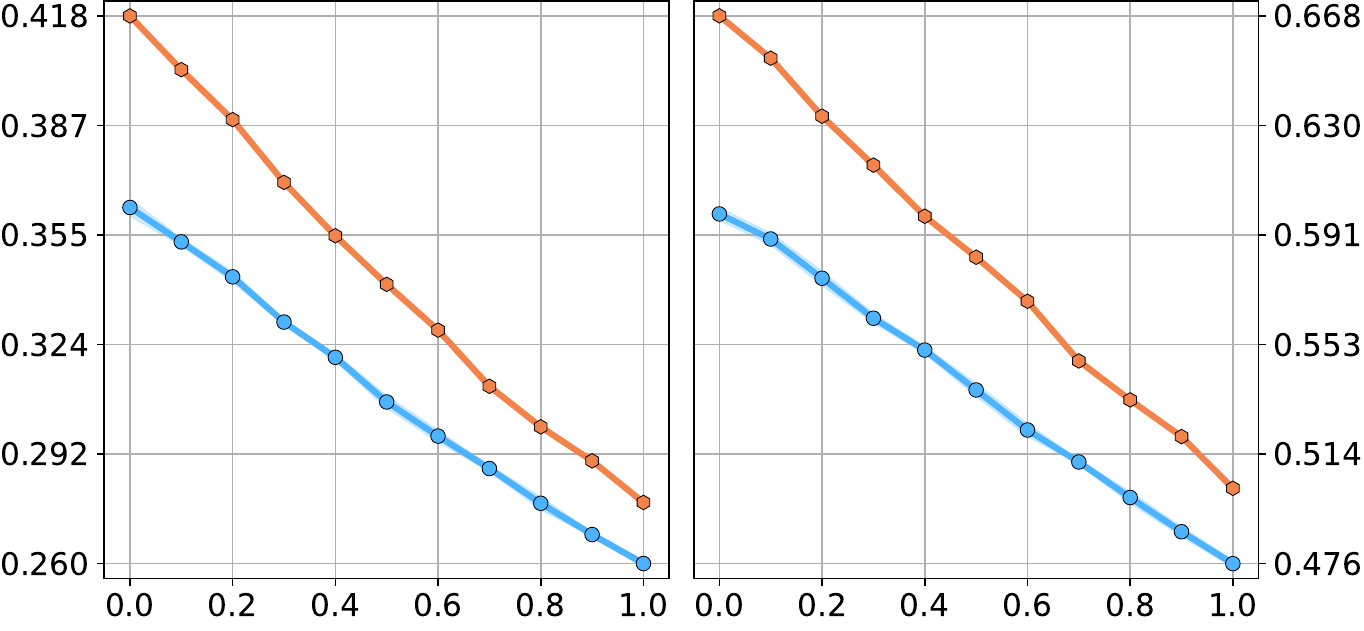}};
        \node (mscocoplot) [anchor=east] at ($(plot.west)+(0,0)$) {\includegraphics[clip, trim={3cm 0.5cm 23.5cm 1.5cm}, height=\figheight]{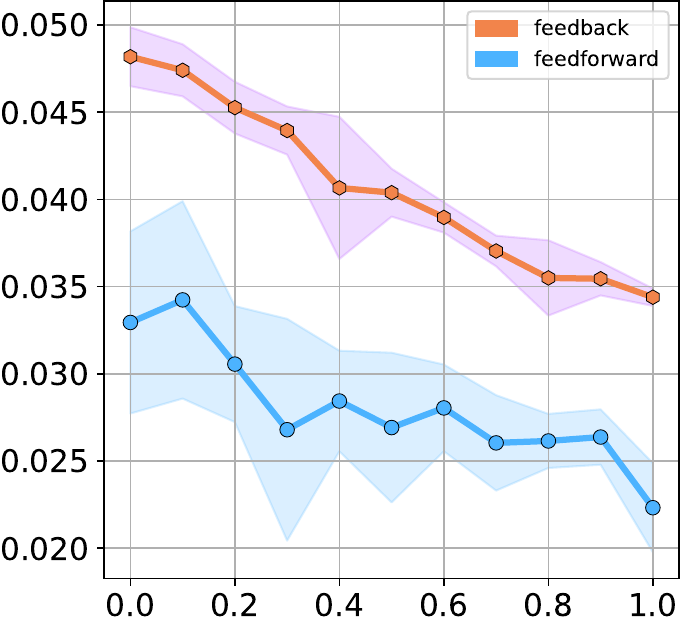}};
        \node (legend1) at ($(mscocoplot.south west)+(\mscocolegendXshift,0)$) {$\sigma$};
        \node (legend2) at ($(plot.south east)+(-\imagenetlegendXshiftL,0)$) {$\sigma$};
        \node (legend3) at ($(plot.south west)+(\imagenetlegendXshiftR,0)$) {$\sigma$};
        \node (mIoU) at ($(mscocoplot.north west)+(\mscocolegendXshift,0.2)$) {mIoU};
        \node (top1acc) at ($(plot.north west)+(\imagenetlegendXshiftR,0.2)$) {Top-$1$ Accuracy};
        \node (top5acc) at ($(plot.north east)+(-\imagenetlegendXshiftL,0.2)$) {Top-$5$ Accuracy};

        \node (A) at ($(mscocoplot.north west)+(0.25,0.3)$) {\sffamily{\textbf{A}}};
        \node (B) at ($(plot.north west)+(0.25,0.3)$) {\sffamily{\textbf{B}}};
    \end{tikzpicture}
    \caption{Performance of feedback versus feedforward in noisy settings. {\sffamily{\textbf{A}}} shows the noise impact of these models in the COCO-Stuff dataset. {\sffamily{\textbf{B}}} shows the noise impact in the Imagenet dataset.}
    \label{fig:noise_imagenet}
\end{figure*}
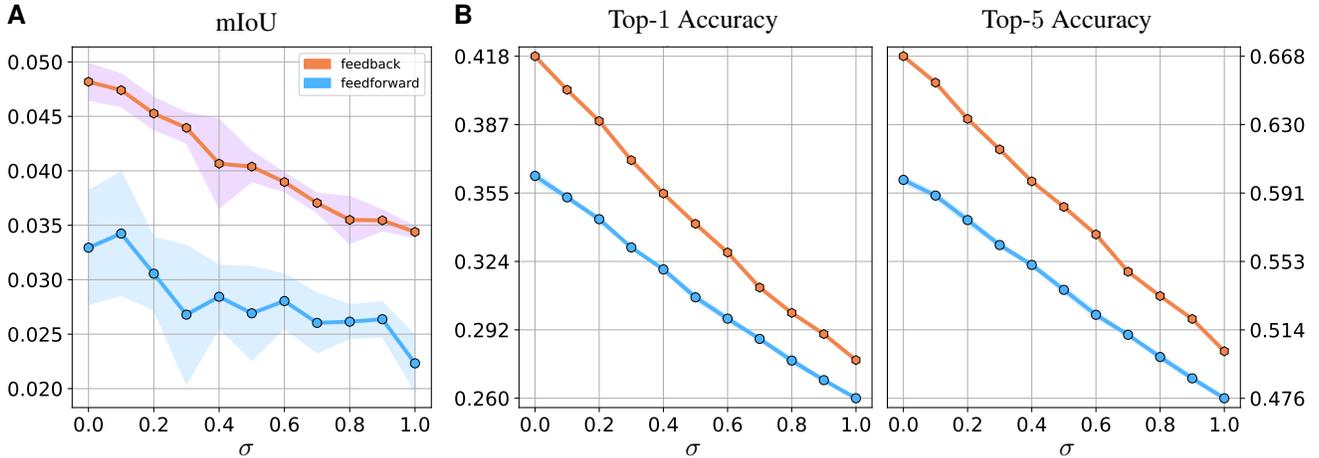

%% file: setting/setting.tex
We assess the robustness and generalization capabilities of our feedback models under two experimental conditions: input noise (denoted by $\sigma$) and number of examples per class (denoted by $D$), as described in Section \ref{section:problem}. We consider three datasets: ImageNet-1k \cite{deng2009imagenet}, used to test both robustness to noise and generalization from limited samples; COCO-Stuff \cite{lin2014microsoft}, used for semantic segmentation under the noise condition; and MedMNIST {\cite{yang2023medmnist}} where we test on the original test sets (Retina, Blood, Derma, Pneumonia, and Breast) and on the RetinaMNIST-C {\cite{di2024medmnist}} to assess performance under different types of noise (brightness down, defocus blur, jpeg compression, pixelate, contrast down, gaussian noise, motion blur, speckle noise). Different neural network architectures are used depending on the task:
\begin{itemize}
    \item For ImageNet classification:
    \begin{itemize}
        \item A ResNet-50 is used for the low-data regime experiments.
        \item A ResNet-18 is used for the noise experiments for faster training.
    \end{itemize}
    \item For semantic segmentation (COCO-Stuff), we use DeepLabV3+ with a ResNet-18 encoder.
    \item For MedMNIST classification, we use a ResNet-50.
\end{itemize}
Input preprocessing differs per setting. In the ImageNet and COCO-Stuff noise experiments inputs are randomly  cropped and resized to $64 \times 64$ (naturally this decreases the expected performance of the models, but speeds up training). Downsampling significantly the resolution for the COCO-Stuff emulates an experiment of noise with lower class density as the number of pixels available is decreased. For the ImageNet generalization experiments, inputs are resized to $224 \times 224$ with random cropping. In the MedMNIST, we perform a random crop followed by a resize to $224\times 224$. We set the time constant \( \tau = 1 \), which controls the rate of decay. 
We unroll the feedback loop for a fixed number of steps \( T = 5 \), so that the final prediction, as described in equation \ref{equation:feedback}, is based on \( \mathbf{h}(5) \). All models are trained using stochastic gradient descent with cross entropy loss, with $0.1$ label smoothing, and batch size $32$. We use the one cycle learning rate scheduler \cite{smith2017cyclical} with initial learning rate of $0.05$, maximum learning rate $1.0$, and final learning rate of $5e-5$. This aggressive schedule enables superconvergence \cite{smith2019super}, which helps expose instabilities in feedback dynamics. In fact, most of the design choices described in Section \ref{section:methods}, such as exponential decay, orthogonalization of $Q$, and normalization by $\frac{1}{Z}$, were necessary to achieve convergence under this setup. 

%% file: results/results.tex
We report results for two main settings, noise and number of examples per class, that address the robustness and generalization abilities, respectively. These are some of the traits that separate computer from human vision, the ability to generalize and robustness to noise. We believe that future model development has to look beyond data and model sizes, and directly address these issues.

\subsection{Robustness}\label{section:robustness}

\input{results/examples/figures/examples_figures}

\input{results/noise/noise}

\subsection{Generalization}\label{section:generalization}

\input{results/examples/examples}

\subsection{Application in medical imaging}\label{section:medical_application}

\input{results/medical/medical}

We evaluate DFMs on multiple \textit{MedMNIST} datasets (Table \ref{tab:medmnist}) and on \textit{RetinaMNIST-C}, a corruption benchmark derived from RetinaMNIST (Table \ref{tab:medmnist_noise}). All models are trained on clean data ($\sigma=0$) and evaluated under eight distribution shifts (blur, compression, pixelation, contrast, additive noise, motion blur, speckle, and brightness changes). Across nearly all corruptions, DFMs outperform a standard ResNet-50 backbone, yielding consistent AUROC gains. In particular, DFMs improve over the feedforward baseline by $+4$pp under \textit{brightness-down} and by $+1.5$pp under \textit{pixelate}, perturbations that substantially degrade the feedforward model. We compare against \textit{Deep Equilibrium Models} (DEQs) \cite{bai2019deep}, which compute gradients at an equilibrium rather than through the entire trajectory; importantly, DEQs lack explicit top-down feedback. We also include \textit{C\mbox{-}RBP} (Stable Recurrent Vision Models) instantiated on a ResNet\mbox{-}50 \textit{backbone} \cite{linsley2005stable}, and \textit{CORNet\mbox{-}R} \cite{kubilius2018cornet}, a biologically motivated \emph{local} recurrence baseline. While both DFMs and DEQs reach equilibrium states, the mechanisms differ: DFMs inject a global top-down signal during unrolling, whereas DEQs do not employ feedback. Overall, DFMs are competitive on most datasets, which we hypothesize stems from the additional top-down information shaping the gradient signal during unrolling. To isolate the role of feedback from mere iteration count, we evaluate a feedback-masked control (DFM-ResNet-50 w. $ v=0$) that zeros the top-down vector for $t>0$ while matching $T$, FLOPs, and parameters. Iterating alone provides some benefit, but using feedback is advantageous on balance. This robustness under covariate shift supports the biological motivation for feedback: refining internal representations when inputs are ambiguous or degraded.

%% file: results/examples/figures/examples_figures.tex
\newcommand{\examplesheight}{7cm}
\newlength{\exampleslegendXshiftL}
\newlength{\exampleslegendXshiftR}
\pgfmathsetlength{\exampleslegendXshiftR}{0.535*\examplesheight}
\pgfmathsetlength{\exampleslegendXshiftL}{0.635*\examplesheight}

\begin{figure*}[t]
    \centering
    \begin{tikzpicture}
        \node (plot) at (0,0) {\includegraphics[clip, trim={0cm 1.25cm 14.5cm 0.25cm}, height=\examplesheight]{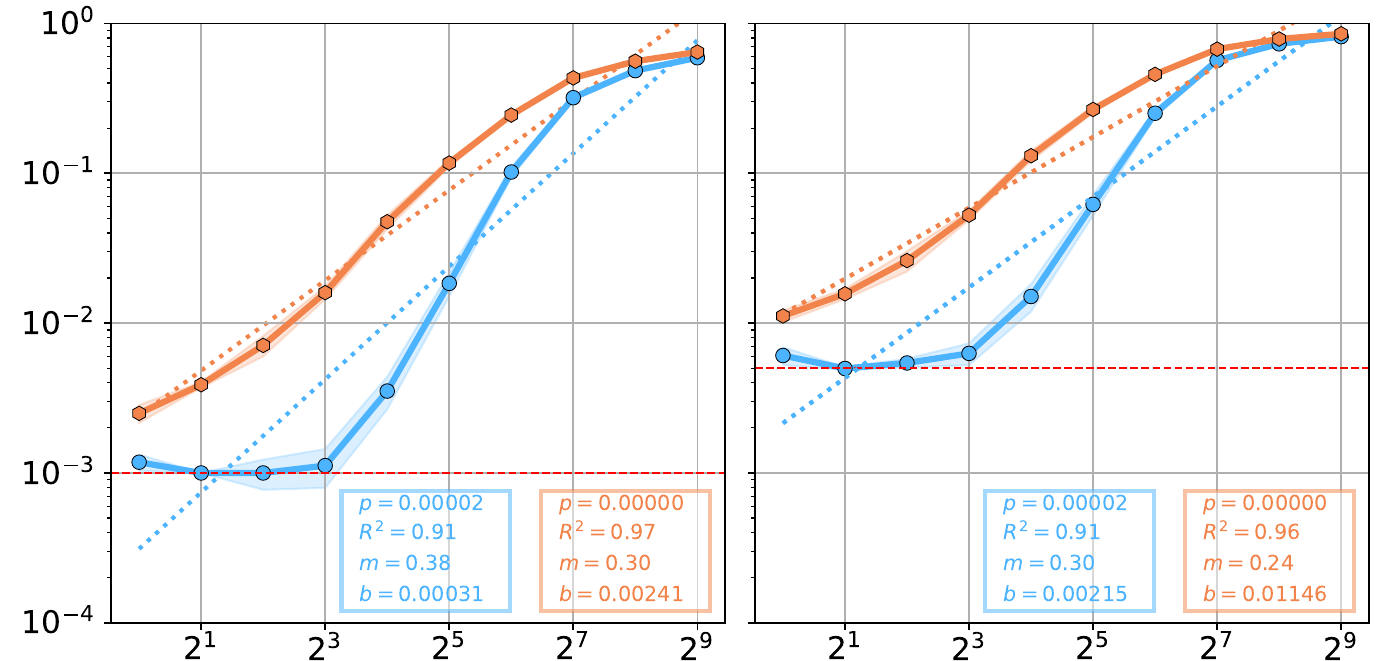}};
        \node (legend1) at ($(plot.south east)+(-\exampleslegendXshiftR,-0.1)$) {$D$};
        \node (legend2) at ($(plot.south west)+(\exampleslegendXshiftL,-0.1)$) {$D$};
        \node (top1acc) at ($(plot.north west)+(\exampleslegendXshiftL,0.)$) {Top-$1$ Accuracy};
        \node (top5acc) at ($(plot.north east)+(-\exampleslegendXshiftR,0.)$) {Top-$5$ Accuracy};
    \end{tikzpicture}
    \caption{Performance of feedback (orange) versus feedforward (cyan). On the left, we plot the top-$1$ accuracy in the ImageNet dataset. On the right, we plot the top-$5$ accuracy. We tested $10$ different values for $D$ (see Section \ref{section:experimental}), which were $\in \{ 1, 2, 4, 8, 16, 32, 64, 128, 256, 512\}$, all represent in base $2$ in the plots. For each $D$ we ran $5$ seeds to plot the error. The horizontal red dotted line represents the performance of a random classifier. We also plot the linear regression (orange and cyan dotted lines) parameters of both feedback and feedforward settings.}
    \label{fig:examples_imagenet}
\end{figure*}
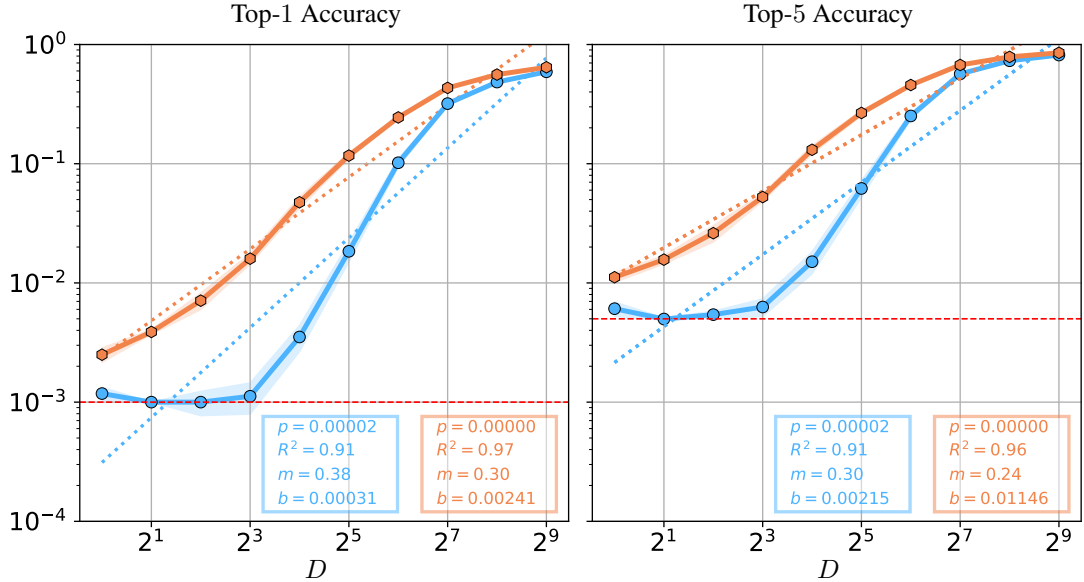

%% file: results/noise/noise.tex
We evaluated the robustness of DFMs against input perturbations by adding Gaussian noise of increasing magnitude to the ImageNet and COCO-Stuff datasets, see Figure \ref{fig:noise_imagenet}. In the COCO-Stuff, DFMs dominate the performance for all levels of noise. 
With no levels of noise, DFMs surpass feedforward models by $+1.4$pp, with DFMs and feedforward claiming $0.047$ and $0.032$ mIoU, respectively. As noise increases the performance gap decreases. At $\sigma=1.$, DFMs had $0.034$, while feedforward $0.022$ mIoU. In the ImageNet, across all noise levels, the feedback model consistently outperformed the feedforward baseline, with its performance curve serving as a strict upper bound. Without noise ($\sigma = 0$), the feedback model achieved an average top-1 accuracy of $41.83\%$, compared to $36.31\%$ for the feedforward model, a notable $+5.52$pp improvement. At the highest noise level tested, accuracy decreased to $27.80\%$ for feedback and $26.04\%$ for feedforward. As expected, the performance gap narrows under extreme noise, since classification approaches random guessing and becomes equally difficult for all models. Despite this narrowing, the feedback model maintained higher top-1 and top-5 accuracy throughout. Notably, the slope of the feedback degradation curve more steep than that of the feedforward model, but DFMs dominated feedforward for all values indicating stronger robustness to noise. All reported values are averaged over 5 random seeds, and standard deviations were negligible, suggesting consistent behavior across runs. These results reinforce the claim that DFMs offer a more noise resilient alternative for object recognition tasks.

%% file: results/examples/examples.tex

We evaluated how well the models generalize under limited supervision by varying the number of training examples per class, $D$, on the ImageNet dataset. Across all conditions, the feedback model consistently outperformed the feedforward baseline in both top-1 and top-5 accuracy. As shown in Figure \ref{fig:examples_imagenet}, both models exhibit a power law scaling behavior with respect to the number of training examples. While the slope of the power law is steeper for the feedforward model, indicating faster gains with more data, the feedback model consistently starts from a higher intercept. This suggests that feedback enables better generalization in extremely low-data regimes. The largest performance gap occurred at $D = 8$, where the DFM achieved a top-1 accuracy of $1.60\%$, compared to only $0.12\%$ for feedforward. Even in the extreme case of $D = 1$, the DFM doubled the performance of the baseline, scoring $0.25\%$ versus $0.12\%$. As the amount of data increases, the performance difference narrows: with $D = 512$, DFM reached $64.43\%$ top-1 accuracy, while feedforward reached $59.00\%$. When trained on the full dataset ($D \approx 1280$), both models surpassed $72\%$ top-1 accuracy. A similar trend holds for top-5 accuracy, with DFM maintaining consistent advantages across the full range of $D$. These results demonstrate that DFMs are particularly advantageous in low-data regimes, offering stronger generalization from very limited supervision.

%% file: results/medical/medical.tex
\begin{table*}
    \centering
    \small{\begin{tabular}{p{4.cm} | p{2.1cm} | p{2.1cm} | p{2.1cm} | p{2.1cm}  | p{2.1cm}}
        \hline
        \hline
        \hfil   & \hfil Retina  & \hfil Blood & \hfil Derma & \hfil Pneumonia & \hfil Breast\\
        \hline
        \hfil ResNet-50 & \hfil $0.707 \pm 0.016$ & \hfil $0.996 \pm 0.001$ & \hfil $\underline{0.919 \pm 0.004}$ & \hfil $0.978\pm0.007$ & \hfil $0.771 \pm 0.054$ \\
        \hfil C-RBP-ResNet-50 \cite{linsley2005stable} & \hfil $0.694 \pm 0.049$ & \hfil $0.998 \pm 0.001$  & \hfil $\mathbf{0.936 \pm 0.008}$ & \hfil $\mathbf{0.990 \pm 0.004}$ & \hfil $0.850 \pm 0.019$ \\
        \hfil CORNet-R \cite{kubilius2018cornet} & \hfil $0.704 \pm 0.011$ & \hfil $0.998 \pm 0.001$  & \hfil $0.914 \pm 0.005$  & \hfil $\underline{0.984 \pm 0.004}$  & \hfil $0.650 \pm 0.084$ \\
        \hfil DEQ-ResNet-50 \cite{bai2019deep} & \hfil $0.693\pm 0.073$ & \hfil $0.995 \pm 0.004$ & \hfil $0.889 \pm 0.058$ & \hfil $0.980 \pm 0.013$ & \hfil $0.634 \pm 0.140$ \\
        \hline
        \hfil DFM-ResNet-50 w. $\mathbf{v}=\mathbf{0}$ & \hfil $\mathbf{0.725 \pm 0.014}$ & \hfil $\mathbf{0.999 \pm 0.000}$ & \hfil $0.915 \pm 0.023$  & \hfil $0.980 \pm 0.005$  & \hfil $\underline{0.862 \pm 0.025}$ \\
        \hfil DFM-ResNet-50 (Ours) & \hfil $\underline{0.722 \pm 0.012}$ & \hfil $\mathbf{0.999 \pm 0.000}$ & \hfil $0.892 \pm 0.020$ & \hfil $0.983 \pm 0.004$ & \hfil $\mathbf{0.881 \pm 0.034}$ \\
        \hline
        \hline
    \end{tabular}}\normalsize
    \caption{AUC of MedMNIST test set. All models were trained with $\sigma=0$ and $D\times L=|\mathcal{D}|$. Results are averaged over $5$ seeds.}
    \label{tab:medmnist}
\end{table*}

\begin{table*}
    \centering
    \small{\begin{tabular}{p{4.cm} | p{1.1cm} | p{1.1cm} | p{1.1cm}  | p{1.1cm}  | p{1.1cm}  | p{1.1cm}  | p{1.1cm}  | p{1.1cm} }
        \hline
        \hline
        \hfil Model & \hfil BD & \hfil DB & \hfil JC & \hfil P & \hfil CD & \hfil GN & \hfil MB & \hfil SN \\
        \hline
        \hfil ResNet-50 & \hfil $0.623$ & \hfil $0.704$ & \hfil $0.705$ & \hfil $0.707$ & \hfil $0.633$ & \hfil $0.701$ & \hfil $0.703$ & \hfil $0.691$ \\
        \hfil C-RBP-ResNet-50 \cite{linsley2005stable} & \hfil $0.633$ & \hfil $\mathbf{0.716}$ & \hfil $\mathbf{0.717}$ & \hfil $\underline{0.717}$ & \hfil $0.631$ & \hfil $\underline{0.713}$ & \hfil $\mathbf{0.715}$ & \hfil $\mathbf{0.713}$ \\
        \hfil CORNet-R \cite{kubilius2018cornet} & \hfil $0.638$ & \hfil $0.700$ & \hfil $0.702$ & \hfil $0.700$ & \hfil $\mathbf{0.680}$ & \hfil $0.703$ & \hfil $0.697$ & \hfil $0.703$ \\
        \hfil DEQ-ResNet-50 \cite{bai2019deep} & \hfil $0.605$ & \hfil $0.708$ & \hfil $0.711$ & \hfil $\underline{0.717}$ & \hfil $0.623$ & \hfil $\mathbf{0.715}$ & \hfil $0.708$ & \hfil $\underline{0.710}$  \\
        \hline
        \hfil DFM-ResNet-50 w. $\mathbf{v}=\mathbf{0}$ & \hfil $\underline{0.643}$ & \hfil $0.712$ & \hfil $0.709$ & \hfil $0.716$ & \hfil $0.626$ & \hfil $0.692$ & \hfil $0.710$ & \hfil $0.696$\\
        \hfil DFM-ResNet-50 (Ours) & \hfil $\mathbf{0.663}$ & \hfil $\underline{0.714}$ & \hfil $\underline{0.714}$ & \hfil $\mathbf{0.722}$ & \hfil $\underline{0.648}$ & \hfil $0.691$ & \hfil $\mathbf{0.715}$ & \hfil $0.707$ \\
        \hline
        \hline
    \end{tabular}}\normalsize
    \caption{AUC of RetinaMNIST in the corrupted instances (MedMNIST-C). The various types of noise: brightness down (\textbf{BD}), defocus blur (\textbf{DB}), jpeg compression (\textbf{JC}), pixelate (\textbf{P}), contrast down (\textbf{CD}), gaussian noise (\textbf{GN}), motion blur (\textbf{MB}), and speckle noise (\textbf{SN}). All models were trained with $\sigma=0$ and $D\times L=|\mathcal{D}|$. Results are averaged over $5$ seeds.}
    \label{tab:medmnist_noise}
\end{table*}

%% file: discussion/discussion.tex
\textbf{Ablation studies.} We performed ablation studies on the role of exponential decay, orthogonality, and the convolution operation proposed. We used the ResNet-18 and trained the models for $70$ epochs. These results are shown in Table {\ref{tab:ablation}}. Removing exponential decay results in a notable drop in performance (from $0.425$ to $0.401$ top-$1$ accuracy), confirming its role in stabilizing the recurrent dynamics by ensuring convergence. Similarly, removing the orthogonalization step during training degrades accuracy ($0.425$ to $0.399$), which aligns with our hypothesis that uncontrolled weight updates can lead to instabilities. Lastly, extending decay to the spatial domain via convolutional exponential decay further improves performance (from $0.425$ to $0.432$), suggesting that spatial dissipation enhances coherence in dense prediction tasks. These results highlight that all three components are beneficial and work in synergy to support robust learning in DFMs. While each mechanism contributes incrementally on its own, their combination yields the most stable and accurate behavior. We found no single component to be redundant, justifying their inclusion in our model.
\begin{table}[ht]
    \centering
    \small{\begin{tabular}{p{4.5cm} | p{1.25cm} | p{1.25cm}}
        \hline
        \hline
          & \hfil \xmark & \hfil \cmark \\
        \hline
        \hfil exponential decay (equation \ref{equation:feedback_decay}) & \hfil $0.401$ & \hfil $\mathbf{0.425}$ \\
        \hfil orthogonality (equation \ref{equation:qr}) & \hfil $0.399$ & \hfil $\mathbf{0.425}$ \\
        \hfil convolution (equation \ref{equation:ode_conv_exp_eig}) & \hfil $0.425$  & \hfil $\mathbf{0.432}$ \\
        \hline
        \hline
    \end{tabular}}\normalsize
    \caption{Ablation study performed with DFM on TinyImageNet (Top-$1$ accuracy). The entries without convolution, with orthogonalization, and with exponential decay are the same model.}
    \label{tab:ablation}
\end{table}

In the COCO-Stuff experiments, we observed an effect of limited class density, while corrupting images with Gaussian noise. All our results validate the hypothesis that DFMs handle noise better and are able to generalize in few shot learning settings. As such, the superiority of DFMs in the segmentation task is due to the latter, as noise is analyzed while the class density is limited.

\begin{figure*}[ht]
    \centering
    \includegraphics[width=1\textwidth]{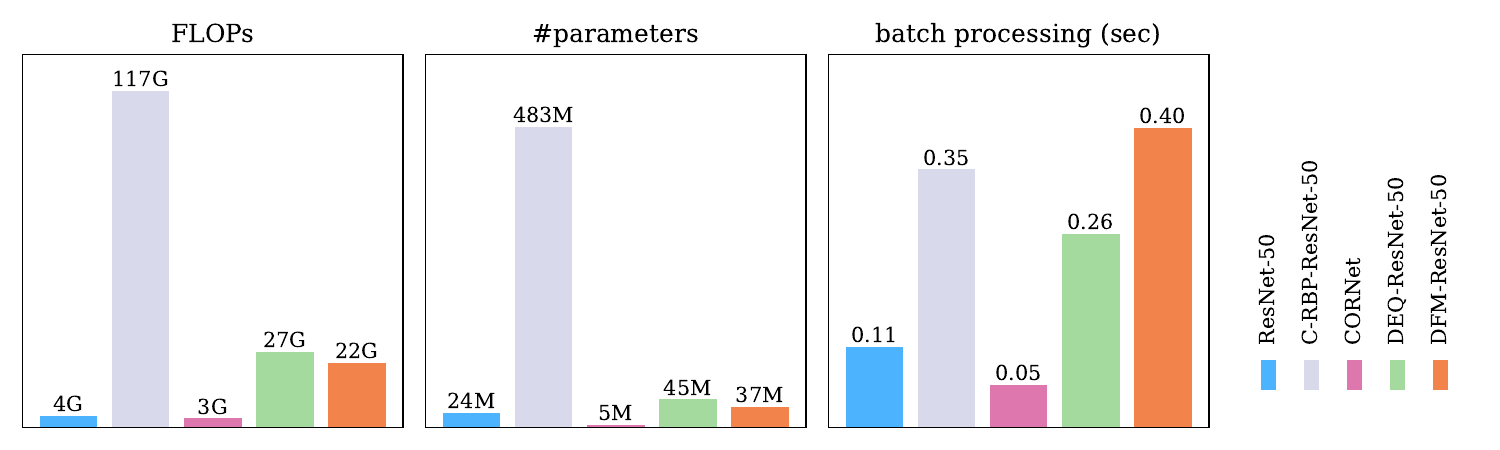}
    \caption{FLOPs, number of parameters, and mean batch processing time for all the baselines considered in this study. Models that implement recurrence, such as C-RBP-ResNet-50, CORnet, and DFM-ResNet-50 are set with $t=5$.}
    \label{fig:dfm_complexity}
\end{figure*}

\noindent\textbf{Feedback is more robust than feedforward.}  
Our experiments show that feedback models consistently outperform their feedforward counterparts in noisy settings, for both classification and segmentation tasks. This suggests that feedback mechanisms are especially well suited for environments with degraded input quality. Moreover, feedback models compute gradients through a recurrent unfolding of the internal state. Because the final representation $\mathbf{h}(T)$ is an accumulation over time, the backward pass implicitly aggregates gradients from all previous steps. The latter is also why we take the loss w.r.t. $\mathbf{y}(T)$. This results in a gradient estimate that is more aligned with the true gradient of the underlying objective, as argued in energy based learning theory \cite{scellier2019equivalence}. In this sense, feedback acts as a form of structured optimization, not just inference.

\noindent\textbf{Feedback generalizes better with fewer examples.}  
In low data regimes, feedback models show stronger generalization ability than feedforward models. On ImageNet, for example, feedback consistently dominates the performance curve across all numbers of training examples per class. Fitting a power law to the accuracy curves, with a statistical significant fit with $p$-values $< 0.01$ and $R^2>0.90$, confirms that both models follow the expected scaling behavior. While feedforward models exhibit a steeper slope, indicating more rapid performance gains as data increases, the feedback models achieve much higher intercepts. In the limit of very few examples, this translates to significantly better performance, sometimes more than 10× higher than feedforward. This reinforces the idea that feedback enables better inductive biases for generalization, particularly in sample efficient learning.

\textbf{Is it the iterations or feedback?} Feedback performs multiple iterative passes, allowing it to refine internal representations over time. However, feedforward models rely on a single pass and lack the capacity to revise their interpretation of ambiguous stimuli. To test whether the advantage comes from the iterative nature or from the feedback, we define $\Delta_i = \mbox{DFM}_i - \left(\mbox{DFM w. }\mathbf{v}=0\right)_i$, where $\mbox{DFM}_i$ are the performances averaged over the seeds of DFM on the $i$th dataset, and test the null hypothesis $H_0:\mathbb{E}[\Delta]\leq0$. We found that DFM was statistically significant against DFM with $\mathbf{v}=0$, using a one-sided paired t-test across datasets, with a $p$-value of $0.015$. DFMs are competitive or superior on \textit{Retina}, \textit{Blood}, \textit{Pneumonia}, and \textit{Breast} (Table \ref{tab:medmnist}), and lead on several realistic corruptions in \textit{RetinaMNIST-C}, notably brightness-down, pixelate, contrast-down, and motion blur (Table \ref{tab:medmnist_noise}). 

\noindent\textbf{Global versus local recurrence.} While prior recurrent architectures such as CORnet \cite{kubilius2018cornet} and C-RBP \cite{linsley2005stable} rely on local recurrence, refining representations layer-by-layer or through Jacobian-constrained updates, our results suggest that global recurrence provides a distinct advantage. By injecting high-level predictions back to the input, DFMs enable top-down corrections that propagate throughout the entire network, rather than remaining confined to shallow feature refinements. Empirically, this yields consistent gains in robustness and generalization on medical imaging benchmarks, even when controlling for iterative depth (see table \ref{tab:medmnist_noise}). In terms of complexity, DFMs are \textbf{not} substantially more expensive than other recurrent approaches: FLOPs, parameter counts, and batch processing times are comparable to CORnet C-RBP (see figure \ref{fig:dfm_complexity}), while remaining within the same order of magnitude as DEQ-based models. Thus, although DFMs incur higher cost than a single-pass feedforward baseline, they offer a favorable trade-off against alternative recurrent designs, achieving stronger robustness to corruptions and distribution shifts, which is particularly valuable in domains such as medical imaging where reliability outweighs raw throughput.

\noindent\textbf{Limitations.} DFMs do come with higher memory and compute costs compared to their feedforward counterparts, and by extension DEQs, mainly because we unroll $\FB$ and $G$ for $T$ iterations during training. However, when comparing against C-RBP-ResNet-50, a suitable model for comparison against ours, we observe that DFMs are $\approx 5$ times more efficient and $\approx 13$ times less parameters, while only observing an increase in training time of $13\%$. 
This is illustrated in Figure \ref{fig:dfm_complexity}. Overall DFMs' trade-offs do come with a clear upside: DFMs offer significantly stronger robustness and generalization. In scenarios where performance under distribution shifts matters, the compute is a worthwhile investment.

%% file: related_work/related_work.tex
\noindent \textbf{Biologically inspired feedback and recurrence.}
Although feedback as defined in this work, explicit high level to low level signal flow, has not been widely explored, there exists a rich literature on models incorporating recurrence. Classical examples include Hopfield networks \cite{hopfield1982neural}, DEQs \cite{bai2019deep}, and Feasibility based Fixed Point Models \cite{heaton2021feasibility}. These approaches typically repeat layers or update a latent state through iterative processes, but do not integrate top down feedback across hierarchical levels. Notably, DEQs perform fixed point iteration until the system reaches equilibrium and then proceeds to compute the gradient at that point. We believe that this process inevitably loses information of the trajectory made during the forward process and its performance is hurt as a consequence. More biologically grounded efforts such as CORnet \cite{kubilius2018cornet} and predictive reconstruction frameworks \cite{alamia2023role, fein2024contextual} propose layer wise recurrence, where each layer predicts its previous and next layers' activations. However, these models rely on local feedback between adjacent layers and lack long range, hierarchical feedback that modulates early representations using high level beliefs. Our approach differs significantly: the entire neural network acts as a recurrent cell, where a global feedback vector is fed back to the input over time. This formulation poses unique challenges, particularly for training stability (see Section \ref{section:orthogonality}), and requires dedicated mechanisms such as exponential decay and orthogonality enforcement.

\noindent \textbf{Stability and robustness in recurrent networks.}
Several works have highlighted the benefits of recurrence for robustness, particularly under noisy or adversarial settings \cite{spoerer2017recurrent}. Stability in such models is often enforced via Jacobian regularization \cite{linsley2005stable} or by limiting the number of unrolled iterations \cite{wang2019recurrent}. In contrast, our model achieves stable convergence through architectural constraints, notably using matrix exponentiation with negative real eigenvalues, without resorting to truncation or gradient clipping.


\noindent \textbf{Connections to recurrent backpropagation.}
Our method also shares similarities with recurrent backpropagation (RBP), where gradients are computed at equilibrium by backpropagating through fixed point dynamics \cite{pineda1987generalization, almeida1990learning, liao2018reviving}. In RBP, gradient convergence depends on the spectral properties of the underlying dynamics, typically requiring matrices with negative real eigenvalues. In our case, we mirror this principle during the forward pass by constructing a computation graph using a matrix $P$ with eigenvalues $\Sigma < 0$ (see Equation \ref{equation:feedback_decay}), ensuring fast convergence to a stable state. Unlike classical RBP, which is designed for constant memory usage, our models store the full trajectory to enable learning through automatic differentiation, at the cost of higher memory usage. This enables training of complex systems that would otherwise be intractable with Jacobian based regularization. Despite the constant memory usage, the number of parameters and FLOPs, reported when using the ResNet-50 as a backbone for a recurrent layer trained with RBP, are the highest among the baselines.

\input{related_work/superdiffusive}
\input{related_work/ff_vs_fb}

%% file: conclusion/conclusion.tex
It is hypothesized that the brain performs decision making by solving a differential equation over time \cite{busemeyer1993decision}. However, most state of the art neural networks used in object recognition and segmentation are static, lacking biologically plausible feedback mechanisms. This limitation makes them brittle in domains such as medical imaging, where data is noisy or scarce. In this work, we introduced DFMs, a novel class of neural architectures that integrate top down feedback into the computational graph. Unlike conventional feedforward models, DFMs update internal representations through a recurrent process that refines predictions over time. This leads to more robust gradients, improved stability, and better generalization, especially under noise and limited data conditions. Our experiments demonstrate that DFMs consistently outperform feedforward baselines in both robustness and data efficiency, confirming the benefits of hierarchical feedback dynamics and making DFMs a natural candidate in medical imaging candidates.


%% file: acknowledgments.tex
\section*{Acknowledgments}

This work was supported by the Center for Responsible AI project with reference C628696807-00454142, financed by the Recovery and Resilience Plan, and INESC-ID pluriannual UIDB/50021/-2020.